%% file: main.tex
\documentclass[runningheads]{llncs}
\usepackage[numbers,sort&compress]{natbib}

\usepackage{graphicx}
\usepackage[dvipsnames]{xcolor}
\usepackage{amsmath}
\usepackage{libertinust1math}
\usepackage{enumitem}
\usepackage{booktabs}

\usepackage{color}
\usepackage{url}

\usepackage{orcidlink}
\newcommand\orcid[1]{\textsuperscript{\orcidlink{#1}}}

\begin{document}

\title{Investigating Retrieval-Augmented Generation Systems on Unanswerable, Uncheatable, Realistic, Multi-hop Queries}
\titlerunning{Unanswerable, Uncheatable, Realistic Multi-hop Queries for RAG Evaluation}

\author{Gabrielle Kaili-May Liu\inst{1}\orcid{0000-0002-0603-1655}\and Bryan Li\inst{2}\orcid{0000-0002-5779-1662}\and \\Arman Cohan\inst{1}\orcid{0000-0002-8954-2724} \and
William Gantt Walden\inst{3,4}\orcid{0000-0001-9931-2861}\and
Eugene Yang\inst{3,4}\orcid{0000-0002-0051-1535}}
\authorrunning{Liu et al.}
\institute{Yale University \and University of Pennsylvania \and Human Language Technology Center of Excellence\and
Johns Hopkins University\\\email{kaili.liu@yale.edu, \{wwalden1,eugene.yang\}@jhu.edu}
}

\maketitle

\begin{abstract}
Real-world use cases often present RAG systems with complex queries for which relevant information is missing from the corpus or is incomplete. In these settings, RAG systems must be able to reject unanswerable, out-of-scope queries and identify failures of retrieval and multi-hop reasoning. Despite this, existing RAG benchmarks rarely reflect realistic task complexity for multi-hop or out-of-scope questions, which often can be cheated via disconnected reasoning (i.e., solved without genuine multi-hop inference) or require only simple factual recall.
This limits the ability for such benchmarks to uncover limitations of existing RAG systems.
To address this gap, we present the first pipeline for automatic, difficulty-controlled creation of un\underline{c}heatable, \underline{r}ealistic, \underline{u}nansw-erable, and \underline{m}ulti-hop \underline{q}uerie\underline{s} (CRUMQs), adaptable to any corpus and domain. We use our pipeline to create CRUMQs over two popular RAG datasets and demonstrate its effectiveness via benchmark experiments on leading retrieval-augmented LLMs. Results show that compared to prior RAG benchmarks, CRUMQs are highly challenging for RAG systems and achieve up to 81.0\% reduction in cheatability scores. More broadly, our pipeline offers a simple way to enhance benchmark difficulty and drive development of more capable RAG systems.
\end{abstract}

\input{1-intro}

\input{2-background}

\input{3-method}

\input{4-exp}
\input{5-results}

\input{6-conclusion}

\begin{credits}
\subsubsection{\ackname} 
This material is based upon work supported by the National Science Foundation Graduate Research Fellowship Program under Grant No. DGE-2139841. Any opinions, findings, and conclusions or recommendations expressed in this material are those of the author(s) and do not necessarily reflect the views of the National Science Foundation.

\subsubsection{\discintname}
The authors have no competing interests to declare that are relevant to the content of this article.

\end{credits}

\vspace{-.35cm}
\begin{table}[h!]
\caption{Sample RAG queries generated via the CRUMQs pipeline.}
    \label{tab:app1} 
    \vspace{-2mm}
\centering
{
\fontsize{7pt}{8pt}\selectfont
\setlength{\tabcolsep}{4pt}
\begin{tabular}{c|p{0.9\linewidth}} \toprule
\# Hops & Question\\\midrule
2 & What historical event shares a name with a protocol used in computer science to achieve agreement among parties?\\\midrule
3 & What is the impact of the estimated average annual loss of ice in Greenland and Antarctica combined on rising sea levels and ocean chemistry?\\\midrule
5 & If the number of iterations in a stochastic linear bandit problem is equal to the number of major crusades, what would be the minimum parameter values that would allow for a `crusade' against suboptimal solutions?\\\bottomrule
\end{tabular}
}
\vspace{-1.17cm}
\end{table}

\begin{table}[h!]
\caption{Complexity and lexical diversity of CRUMQs versus prior RAG benchmarks.}
    \label{tab:app2} 
    \vspace{-2mm}
\centering
{
\fontsize{7pt}{8pt}\selectfont
\setlength{\tabcolsep}{5pt}
\begin{tabular}{l|cccccc} \toprule
& Avg Gold  & Avg \#	& Unique \#  & \# Verb-Noun \\
& Ans Len 	& Hops	    & Verbs	     & Uniq.  Pairs \\
\midrule
UAEval4RAG \cite{uaeval4rag} &	67.29	&1.00	&155	&717	\\
MultihopRAG \cite{multihoprag} &	1.32	&2.38	&109	&332	\\
CRUMQs (Ours)&	20.04	&\textbf{2.74}	&\textbf{225}	&\textbf{826}\\\bottomrule
\end{tabular}
}
\vspace{-1.16cm}
\end{table}

\begin{table}[h!]
\caption{Comparative distribution of the number of hops per query.}
    \label{tab:app3} 
    \vspace{-2mm}
\centering
{
\fontsize{7pt}{8pt}\selectfont
\setlength{\tabcolsep}{6pt}
\begin{tabular}{l|ccccccc} \toprule
 &  $\leq$1 & 2 & 3 & 4 & 5 & 6 & 7\\\midrule
MultiHop-RAG \cite{multihoprag} &	11.80\% &	42.20\%&	30.50\%&	15.60\%&	0.00\%&	0.00\%&	0.00\%\\
CRUMQs (Ours) & 11.50\% & 31.90\% & 31.60\% & 21.60\% & 3.10\% & 0.20\% & 0.10\%\\\bottomrule
\end{tabular}
}
\vspace{-0.3cm}
\end{table}

\bibliographystyle{splncs04}
\bibliography{main_biblio, no_cite}

\end{document}

%% file: 1-intro.tex
\section{Introduction}

Retrieval Augmented Generation (RAG) \cite{lewis2020retrieval,asai} is a powerful approach for many NLP tasks, enabling LLMs to respond to diverse user requests by leveraging an external document collection. 
While RAG is highly effective at increasing model credibility \cite{gao2023}, mitigating hallucinations, and improving response quality \cite{pmlr-v162-borgeaud22a}, there remains a need to better understand how such systems handle complex, multi-part queries when available information from the corpus is insufficient. In particular, RAG systems must be able to appropriately reject unanswerable queries (i.e., those for which no relevant information is present in the corpus) and localize retrieval or reasoning failures when responding to multi-hop requests \cite{multihoprag,uaeval4rag}. These capabilities are crucial for reliable deployment of RAG systems in high-stakes domains where information is often missing or incomplete.

Existing RAG benchmarks \cite{rageval,ragbench} rarely evaluate systems’ ability to handle \textit{realistic} multi-hop or unanswerable queries. Multi-hop RAG benchmarks \cite{multihoprag,mhtsrag} focus on synthetic tasks or can be cheated via disconnected reasoning \cite{dire,musique} wherein shortcuts can be exploited to achieve correct answers, %
while unanswerable RAG benchmarks \cite{uaeval4rag} remain limited to simplistic factual recall settings. 

To address these deficiencies, we present the first pipeline for automatic generation of un\underline{c}heatable, \underline{r}ealistic, \underline{u}nanswerable, \underline{m}ultihop \underline{q}uerie\underline{s} (CRUMQs), which are robust against reasoning shortcuts, target content beyond retrieval-augmented LLMs' training data cutoff dates, and can be tailored to any document corpus. We leverage recent insights in synthetic data generation to ensure coverage of diverse task types and complexity levels, with benchmark difficulty easily controllable via the distribution of in- vs.\ out-of-scope hops per question. 

We use our pipeline to create CRUMQs over two popular RAG datasets and showcase its efficacy through experiments on leading retrieval-augmented LLMs. Analysis reveals that CRUMQs pose notable difficulty even for RAG systems employing SOTA models such as GPT-5 \cite{gpt5}.
We further show that CRUMQs are significantly less cheatable via disconnected reasoning than prior multi-hop RAG benchmarks, achieving up to an 81.0\% decrease in cheatability. Overall, our work contributes to a better understanding of RAG systems’ limitations in handling unanswerable queries. Beyond driving the development of stronger and more capable RAG systems, our pipeline opens the door to automatically increasing the difficulty of existing datasets, addressing the challenge of benchmark longevity.

%% file: 2-background.tex
\section{Related Work}
A few studies have examined RAG system performance on queries which either require multi-hop reasoning or are beyond the scope of the relevant document collection \cite{ragsynth}. 
However, these RAG benchmarks are restricted to singular domains and reflect low task complexity: multi-hop queries are fully answerable given the associated corpus and involve $\leq$4 hops, while the out-of-scope queries generally target short factual recall tasks.
For instance, MultiHop-RAG \cite{multihoprag} targets the news domain, with queries based on 2-4 document chunks, but nearly 90\% of questions can be solved by GPT-4---reflecting low difficulty and reduced benchmark value. 
MHTS \cite{mhtsrag} generates
difficulty-controllable multi-hop RAG queries, but QA pairs are created over a \emph{single} novel to evaluate a \emph{single} RAG system, limiting generalizability. 
On the other hand, UAEval4RAG \cite{uaeval4rag} 
presents a framework to create out-of-database and inherently unanswerable RAG requests, yet resulting queries exhibit low complexity, lack difficulty modulation, and may overlap with models' parametric knowledge due to limited data verification. 
Beyond these, common multi-hop benchmarks used for RAG tend to suffer from disconnected reasoning, not involve unanswerability, or reflect limited response formats (e.g., only short entities) \cite{hqa,crag,2wikimultihopqa,squad20,musique}. 
Other QA datasets present both answerable and unanswerable queries, but these adopt narrow task formulations, are not multi-hop, and/or do not involve retrieval \cite{uaqfact,clapnq,factguard,confusionqa,gonzalez-torres-etal-2024-automated,umwp}. In contrast, we present the first pipeline for creating queries tailored to a given corpus that are \emph{both} unanswerable \emph{and} multi-hop---\emph{and} of realistic complexity.

%% file: 3-method.tex
\section{Method}

\begin{figure}[t]
    \centering
    \includegraphics[height=0.55in]{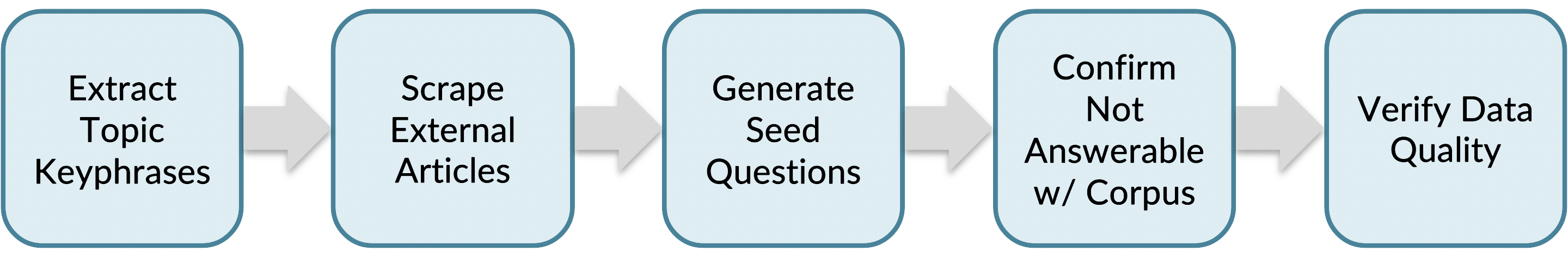}\vspace{-2mm}
    \caption{Overview of the CRUMQs generation pipeline.}
    \label{steps}
    \vspace{-4mm}
\end{figure}

We follow the pipeline shown in Fig.\ \ref{steps} to generate CRUMQs. We begin by extracting relevant topic keyphrases over the provided document collection. In this work, we assume the collection is a RAG corpus $D$ with associated information-seeking requests and gold relevant documents from $D$ per request. 
However, our pipeline may be generalized by using synthetic requests or topic modeling.

\textbf{Step I.} Topics are extracted via two simple steps. A frontier LLM is few-shot prompted to extract short keyphrases from each information-seeking request.\footnote{Our code and prompts are available at \url{https://github.com/pybeebee/CRUMQs}.} To obtain finer-grained topics, we then pair each initial topic with a gold document for the request and employ a second few-shot prompt to extract document-grounded topics. To ensure topics are sufficiently distinct and have good coverage of the corpus, we perform a deduplication step via embedding similarity.\footnote{Other similarly effective deduplication methods may be used, such as a LLM call.} We use the \texttt{BAAI/bge-large-en-v1.5} embedder with a similarity threshold of 0.95.

\textbf{Step II.} To collect relevant information that is likely beyond the given corpus and training data cutoff date for retrieval-augmented LLMs, we next crawl the $N_{e}$ most related recent articles for each topic from the external sources Google News \cite{news}, bioRxiv \cite{bio}, chemRxiv \cite{chem}, medRxiv \cite{med}, arXiv \cite{arxiv}, and PubMed \cite{pubmed}. 

\textbf{Step III.} The externally sourced articles are then used to generate queries that are either \textit{fully} unanswerable (relevant information is not found in $D$, only in the externally sourced documents) or \textit{partially} unanswerable (relevant information is in $D$, but at least one key fact or detail required to provide a complete and correct answer is absent).

To do so, we first split each gold article and each external article into 1,024-token chunks via LangChain \cite{langchain}, tracking for each chunk its source, URL, associated topic keyphrase, and associated request. Chunks are filtered for relevance to the topic and request via a binary LLM judgment. We then construct the set of all possible groups of 2-6 chunks such that between 1-5 externally sourced chunks are in each group. As this set may be overly large in practice, we place a limit $N_c$ on the number of contexts that are created for each total number of chunks and each ratio of external:gold chunks. 

To obtain seed queries, we prompt a strong generator LLM to systematically create up to 10 multi-hop QA pairs using each multi-chunk context,\footnote{While some works explore QA generation over extracted claims \cite{multihoprag} or entity-relation triplets \cite{factcg} from chunks, we found direct use of chunks to be more fruitful.} 
making sure information is leveraged across chunk boundaries. For fully unanswerable queries we require all chunks in a context to be externally sourced; for partially unanswerable queries we require at least one gold chunk. Systematicity and diversity of resulting task formulations are ensured by adapting prompts from prior multi-document QA pipelines \cite{mdcure,longcite}.

\textbf{Step IV.} The unanswerability of each seed question is verified as in UAEval4RAG \cite{uaeval4rag}: each question is used to retrieve the top 10 relevant chunks from the original corpus $D$, and LLM judgment is used to verify these chunks cannot answer the question. QA pairs that pass this verification are retained.

\textbf{Step V.} The data is finally filtered to ensure only truly multi-hop and high-quality queries remain. Following \cite{cot}, we first annotate each QA pair with a synthetic chain-of-thought (CoT) explanation of the answer in the oracle setting (i.e., assuming both gold and externally sourced documents are available), and record the number of hops required to solve the question. Queries that do not adhere to the intended hop count are discarded.\footnote{We retain a selection of single-hop queries in \S\ref{exps} for comparability to prior works.} We then utilize a strong LLM to assess each unanswerable QA pair in the oracle setting according to the following criteria \cite{DRAGON,multihoprag}: context necessity, context sufficiency, answer correctness, answer uniqueness. Contextual criteria are additionally assessed assuming only the given corpus is available. We utilize the same scoring scale (Likert, 0-2) as \cite{DRAGON}: only QA pairs which receive a score of at least 1 for all 6 criteria are retained. A subset of examples is manually reviewed to validate the LLM verification results. %

%% file: 4-exp.tex
\section{Experimental Setup} \label{exps}

We demonstrate the efficacy of our pipeline in creating 
unanswerable, uncheatable, realistic, and multi-hop RAG queries
by comparing against established RAG benchmarks. We consider UAEval4RAG \cite{uaeval4rag} and MultiHop-RAG \cite{multihoprag} as leading baselines, as they are the most recent works to perform comprehensive unanswerable and multi-hop RAG evaluation, respectively.

\textit{Datasets.} 
Using our pipeline, we first generate CRUMQs over NeuCLIR \cite{neuclir} and TREC RAG 2025 \cite{trecrag},
two popular RAG datasets which each consist of a large document collection with associated user requests. We focus on English texts, set $N_e=200$ and $N_c=50$, and use Llama3.3-70B-Instruct \cite{llama3} as the generation and verification model to balance costs and quality.\footnote{As we show in \S\ref{sec:results}, creating effective CRUMQs does depend on proprietary models.} This leads to a total of 3,048 CRUMQs. We next use UAEval4RAG to generate baseline 
out-of-database RAG 
queries over the same two datasets, 
leading to 7,559 out-of-database queries. Finally, we utilize the MultiHop-RAG dataset as-is, which consists of 2,556 multi-hop RAG queries and their associated gold contexts. Comparative dataset statistics as well as sample CRUMQs are provided in \S\ref{sec:results}. 

\textit{Unanswerability Evaluation.} 
To verify the unanswerability value of our pipeline, we adopt the same experimental setup as in \cite{uaeval4rag} and benchmark the same four leading RAG systems from \cite{uaeval4rag} on the CRUMQs and UAEval4RAG queries, downsampled to 3,048 examples for fair comparison. We assume no access to external documents and use LlamaIndex \cite{Liu_LlamaIndex_2022} for RAG system implementation.
We additionally follow \cite{uaeval4rag} to evaluate RAG system performance when using leading proprietary LLMs Gemini-2.5-Pro \cite{gemini2.5}, GPT-4o, and GPT-5 for generation.

\textit{Cheatability Evaluation.} 
To demonstrate the robustness of our queries against reasoning shortcuts, we compare our CRUMQs (downsampled to 2,556 examples) against MultiHop-RAG \cite{multihoprag}. Following \cite{dire}, we first obtain LLM predictions for each dataset in the oracle setting on the tasks of answer prediction and paragraph-level support identification.
We use Llama3.1-8B-Instruct, Llama3.3-70B-Instruct, GPT-4o, and Gemini-2.5-Pro to ensure coverage of diverse model types and sizes. We then repeat the same experiment on the DiRe probe \cite{dire} of each dataset, which is a dataset transformation designed to measure cheatability by gauging the extent to which correct answers can be achieved via reasoning over disjointed input contexts. Comparison of probe to non-probe performance serves as a model-agnostic estimate of cheatability; additional details are in \cite{dire}. 

\textit{Metrics.} 
For unanswerability evaluation, we adopt the metrics of acceptable ratio, unanswered ratio, and ask-for-clarification ratio from \cite{uaeval4rag}. We additionally score accuracy by running LLM judgments of semantic equivalence between target and predicted answers (Gemini-2.0-Flash prompted as in \cite{mdcure}). For cheatability evaluation, we compute the average F1 score for each model$\times$dataset$\times$task setting as in \cite{musique,dire}. The cheatability of each dataset is then measured as the ratio of F1 scores in the probe vs. non-probe settings, which represents the percentage of model performance attributable to disconnected reasoning.

%% file: 5-results.tex
\section{Results and Analysis} \label{sec:results}
\begin{figure}[t]
\centering
\begin{minipage}[t]{0.45\textwidth}
    \centering
    \includegraphics[height=1.85in]{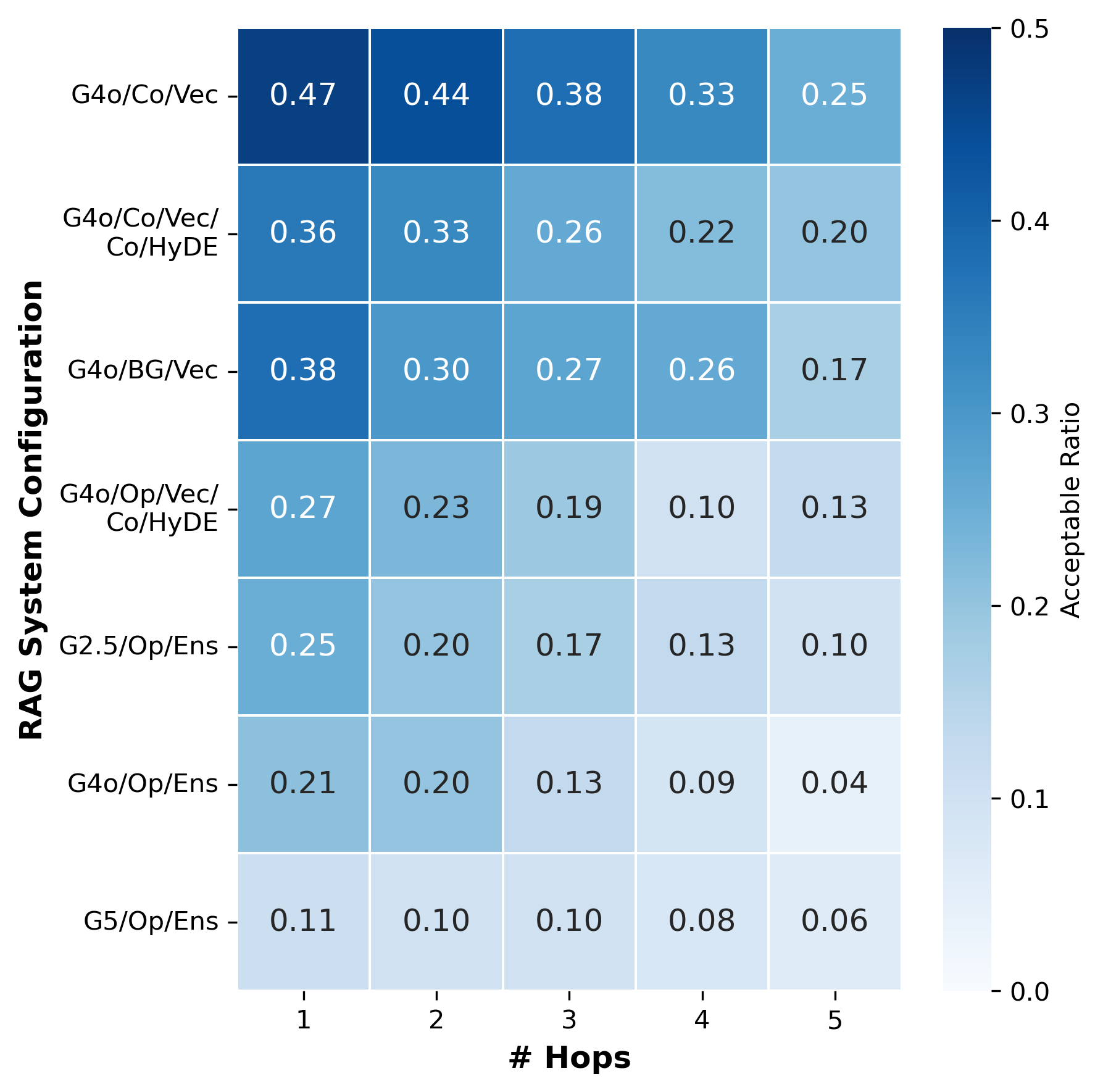}\vspace{-3mm}
    \caption{Acceptable ratios of RAG systems on CRUMQs across hop counts. Performance drops with more hops.}
    \label{fig:hops}\vspace{-4mm}
\end{minipage}\hfill 
\begin{minipage}[t]{.52\textwidth}
    \centering
    \includegraphics[height=1.85in]{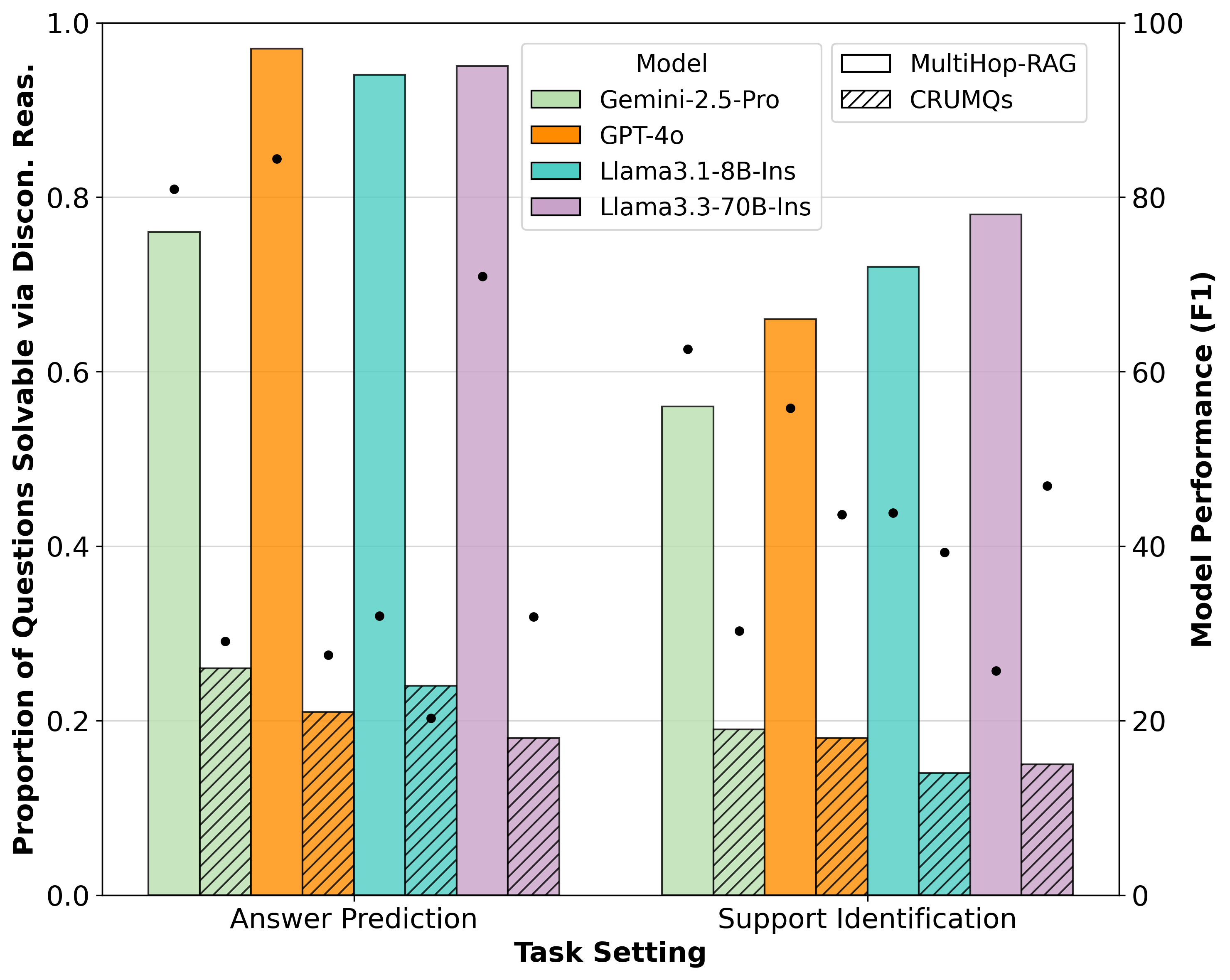}\vspace{-3mm}
    \caption{DiRe F1 score ratios $(\downarrow)$ across benchmarks and tasks. Black points denote accuracy per model per task (values on right axis).}
    \label{fig:cu}\vspace{-4mm}
\end{minipage}
\end{figure}

\begin{table}[t]
\caption{Performance of leading RAG systems \cite{uaeval4rag} using different embedding, retrieval, reranking, and rewriting methods with GPT-4o generation on CRUMQs vs. UAEval4RAG queries. \textsuperscript{\textdagger} denotes Holm-Bonferroni corrected significance ($p<0.05$).}
    \label{tab:ue1}
\centering 
{ %
\fontsize{7pt}{8pt}\selectfont
\setlength{\tabcolsep}{3pt}
\begin{tabular}{l|llll|cccc} \toprule
Dataset	&	Embedding	&	Retrieval	&	Reranker	&	Rewriting	&	Accep. $\uparrow$	&	Unans. $\uparrow$	&	Clar. $\uparrow$	&	Acc. $\uparrow$\\\midrule
UAEval4RAG	&	Cohere	&	Vector	&	None	&	None	&	0.43	&	0.37	&	0.03	&	0.34\\
	&	Cohere	&	Vector	&	Cohere	&	HyDE	&	\textbf{0.50}	&	\textbf{0.94}	&	0.05	&	0.01\\
	&	BGE	&	Vector	&	Cohere	&	None	&	\textbf{0.50}	&	\textbf{0.94}	&	0.05	&	0.01\\
	&	OpenAI	&	Vector	&	Cohere	&	HyDE	&	0.42	&	0.45	&	0.05	&	0.28\\\midrule
CRUMQs	&	Cohere	&	Vector	&	None	&	None	&	\textbf{0.34}\textsuperscript{\textdagger}	&	0.48\textsuperscript{\textdagger}	&	0.06\textsuperscript{\textdagger}	&	0.18\textsuperscript{\textdagger}\\
	&	Cohere	&	Vector	&	Cohere	&	HyDE	&	0.29\textsuperscript{\textdagger}	&	0.79\textsuperscript{\textdagger}	&	0.21\textsuperscript{\textdagger}	&	0.00\textsuperscript{\textdagger}\\
	&	BGE	&	Vector	&	Cohere	&	None	&	0.29\textsuperscript{\textdagger}	&	\textbf{0.80}\textsuperscript{\textdagger}	&	0.20\textsuperscript{\textdagger}	&	0.00\textsuperscript{\textdagger}\\
	&	OpenAI	&	Vector	&	Cohere	&	HyDE	&	0.23\textsuperscript{\textdagger}	&	0.66\textsuperscript{\textdagger}	&	0.05\textsuperscript{\textdagger}	&	0.06\textsuperscript{\textdagger}\\\bottomrule
    \end{tabular}}
     \vspace{-3mm}
\end{table}
We report unanswerability evaluation results in Tables \ref{tab:ue1} and \ref{tab:ue2}. Consistent with the findings in \cite{uaeval4rag}, we observe in Table \ref{tab:ue1} that no single system achieves the best performance on both CRUMQs and UAEval4RAG queries, with different 
system configurations
able to yield similar results.
Importantly, \textit{CRUMQs are harder than UAEval4RAG queries.} All systems respond less acceptably to CRUMQs than UAEval4RAG queries. Moreover, CRUMQs pose notable difficulty for leading proprietary LLM-based RAG systems, 
which provide much fewer acceptable responses for CRUMQs versus UAEval4RAG queries 
(Table \ref{tab:ue2}). While queries from both systems occasionally capture parametric knowledge (i.e., answerable queries are produced, indicated by nonzero accuracy), this proportion is significantly lower for CRUMQs than for UAEval4RAG queries. Finally, CRUMQs with greater hop counts lead to consistently lower acceptable ratios across systems (Fig. \ref{fig:hops}), indicating easily controllable difficulty level for queries. Overall, our pipeline enables creation of harder out-of-database queries with controllable difficulty, which can be used to more effectively differentiate RAG systems.

\begin{table}[t]
\caption{Performance of leading RAG systems \cite{uaeval4rag} using different proprietary models for generation on CRUMQs vs. UAEval4RAG queries. As in \cite{uaeval4rag}, no reranking or rewriting is used. \textsuperscript{\textdagger} denotes Holm-Bonferroni corrected significance ($p<0.05$).}
    \label{tab:ue2} 
\centering
{
\fontsize{8pt}{9.5pt}
\fontsize{7pt}{8pt}\selectfont
\setlength{\tabcolsep}{4pt}
\begin{tabular}{l|lll|cccc} \toprule
Dataset	&	LLM	&	Embedding	&	Retrieval	&	Accep. $\uparrow$	&	Unans. $\uparrow$	&	Clar. $\uparrow$	&	Acc. $\uparrow$ \\\midrule
UAEval4RAG	&	Gemini-2.5-Pro	&	OpenAI	&	Ensemble	&		0.48	&	0.51	&	0.00	&	0.39\\
	&	GPT-5	&	OpenAI	&	Ensemble	&		0.28	&	0.31	&	0.07	&	0.33\\
	&	GPT-4o	&	OpenAI	&	Ensemble	&		\textbf{0.67}	&	0.01	&	0.43	&	0.26\\\midrule
CRUMQs	&	Gemini-2.5-Pro	&	OpenAI	&	Ensemble	&		0.23\textsuperscript{\textdagger}	&	0.23\textsuperscript{\textdagger}	&	0.03\textsuperscript{\textdagger}	&	0.19\textsuperscript{\textdagger}\\
	&	GPT-5	&	OpenAI	&	Ensemble	&		\textbf{0.28}\textsuperscript{\textdagger}	&	0.03\textsuperscript{\textdagger}	&	0.35\textsuperscript{\textdagger}	&	0.18\textsuperscript{\textdagger} \\
	&	GPT-4o	&	OpenAI	&	Ensemble	&		0.24\textsuperscript{\textdagger}	&	0.28\textsuperscript{\textdagger}	&	0.01\textsuperscript{\textdagger}	&	0.12\textsuperscript{\textdagger}\\\bottomrule
    \end{tabular}
}
\end{table}

Compared to prior multi-hop RAG benchmarks, CRUMQs are significantly more difficult and much less cheatable, with all comparisons significant after Holm–Bonferroni correction ($p<0.05$). As indicated by the black points in Fig. \ref{fig:cu}, we observe that CRUMQs pose a much greater challenge for leading LLMs than MultiHop-RAG across all tasks, with models achieving answer prediction scores up to only 31.9 versus up to 84.4 for MultiHop-RAG. Moreover, up to 96.7\% of queries in MultiHop-RAG can be answered via disconnected reasoning, compared to only up to 23.6\% of our CRUMQs. A similar trend is observed for the support identification task.
These results confirm the efficacy of our pipeline for creation of challenging multi-hop RAG queries.\footnote{
We provide sample queries generated via the CRUMQs pipeline in Table \ref{tab:app1} and comparative dataset statistics in Tables \ref{tab:app2} and \ref{tab:app3}.
}

%% file: 6-conclusion.tex
\section{Conclusion and Future Work}
In this work, we introduced an adaptable framework that is the first of its kind for creating highly challenging, difficulty-controllable, unanswerable and multi-hop RAG queries that cannot be easily cheated. Experiments demonstrate that our pipeline effectively addresses deficiencies of existing RAG benchmarks. In addition to advancing the development of stronger RAG systems, our work paves the way toward automatically increasing benchmark difficulty through data augmentation. Future work may explore the performance of multi-hop-oriented RAG systems \cite{mmr,scmrag,hoprag,cg-rag} on CRUMQs, along with analysis of systems’ ability to localize sources of unanswerability and signal uncertainty in such settings.